\def\year{2018}\relax
\documentclass[letterpaper]{article} 
\usepackage{aaai18}  
\usepackage{microtype}
\usepackage{graphicx}
\usepackage{subcaption}
\usepackage{booktabs} 
\usepackage{amssymb}
\usepackage{slashbox}

\usepackage{hyperref}

\usepackage[noend]{algorithmic}

\usepackage{times}
\usepackage[table]{xcolor}
\usepackage{collcell}
\usepackage{hhline}
\usepackage{pgf}
\usepackage{multirow}
\usepackage{float}
\usepackage{algorithm}
\usepackage{amsmath,amsthm}
\usepackage{amssymb}
\usepackage{physics}
\usepackage{graphicx}     

\usepackage[utf8]{inputenc} 
\usepackage[T1]{fontenc}    
\usepackage{hyperref}       
\usepackage{url}            
\usepackage{booktabs}       
\usepackage{amsfonts}       
\usepackage{nicefrac}       
\usepackage{microtype}      

\usepackage{xcolor,colortbl}
\usepackage{color}

\def\colorModel{hsb} 

\newcommand\ColCell[1]{
  \pgfmathparse{#1<11?1:0}  
    \ifnum\pgfmathresult=0\relax\color{white}\fi
  \pgfmathsetmacro\compA{0}      
  \pgfmathsetmacro\compB{#1/20} 
  \pgfmathsetmacro\compC{1}      
  \edef\x{\noexpand\centering\noexpand\cellcolor[\colorModel]{\compA,\compB,\compC}}\x #1
  } 
\newcolumntype{E}{>{\collectcell\ColCell}m{0.4cm}<{\endcollectcell}}  

\title{Make Hawkes Processes Explainable by Decomposing Self-Triggering Kernels}

\author{
  Rafael de Lima \and Jaesik Choi \\
  Department of Computer Science Engineering \\
  Ulsan National Institute of Science and Technology
  Ulsan, Republic of Korea \\
  \{rafael, jaesik\}@unist.ac.kr
}

\begin{document}

\maketitle





\begin{abstract}

In time series data, Hawkes Processes model mutual-excitation between temporal events when the arrival of an event makes future events more likely to happen. Identification of such temporal covariance can reveal the underlying structure to predict future events better. In this paper, we present a new framework to decompose complex covariance structure with a composition of multiple basic self-triggering kernels. 
Our composition scheme decomposes the empirical covariance matrix into the sum or the product of base kernels which are easily interpretable. 
Here, we present the first multiplicative kernel composition methods for Hawkes Processes.  We demonstrate that the new automatic kernel decomposition procedure outperforms the existing methods on the prediction of discrete events in real-world data.
 
 \end{abstract}

\section{Introduction}

Hawkes Processes (HPs) \cite{AH71} model self-exciting behavior, i.e., when the arrival of one event makes future events more likely to happen. This type of behavior has been observed in various domains, such as earthquakes, financial markets, web traffic patterns, crime rates \cite{SL14,GM12} and social media \cite{QZ15}.

As an example, in high-frequency finance, buyers and sellers of stocks demonstrate herding behavior \cite{TL11,BM16}. After the main earthquake, several aftershocks follow according to a time-clustered pattern \cite{YO99}. In web data, hyperlink proliferation across pages exhibit self- and mutual-excitation \cite{JE16}. In criminology, gang-related retaliatory crime patterns are grouped in time \cite{SL14}. In social media, the `infectiousness' of posts can be shown to be modeled through a self-excitement and mutual-excitement assumption \cite{QZ15}.

In HPs analysis, parametric kernels capture intra-domain typical behaviors: quick time-decaying exponential excitation in the case of finance and web data \cite{BM12,JE16}; slower power-law decay in earthquake-related data \cite{YO99}; and periodicity-inducing sinusoidal kernel in TV-watching data \cite{HX16}.

When an appropriate kernel is selected, the kernel parameters are fitted to predict future events. The parameters may be fitted to the data through the gradient descent (GD) method over a likelihood function penalized by a regularization criterion (e.g., Akaike Information Criterion) on the number of parameters \cite{TO79}. Another method of kernel estimation is through the use of the power spectrum of the second order statistics of the process: covariance density and normalized covariance \cite{AH71}. These are well defined when the self-triggering function induces what is called \textit{stationary behaviour}.

However, kernel selection in HPs analysis is a challenging problem, since an appropriate kernel should be manually selected in practice. In this paper, we present a kernel structure search algorithm for HPs. Given base kernels, our algorithm finds the best fitting one, considering composition (sum and product) of base kernels. For verifying the stationarity property of each composite kernel, we also derived analytical expressions for the stationarity conditions. To our best knowledge, our method is the first multi-type kernel composition framework for HPs.

The main steps of the automatic framework, which will be thoroughly explained in the following sections, are discretized kernel estimation and greedy search in kernel composition space.
\begin{figure}[th!]
\centering
\includegraphics[width=0.9\linewidth]{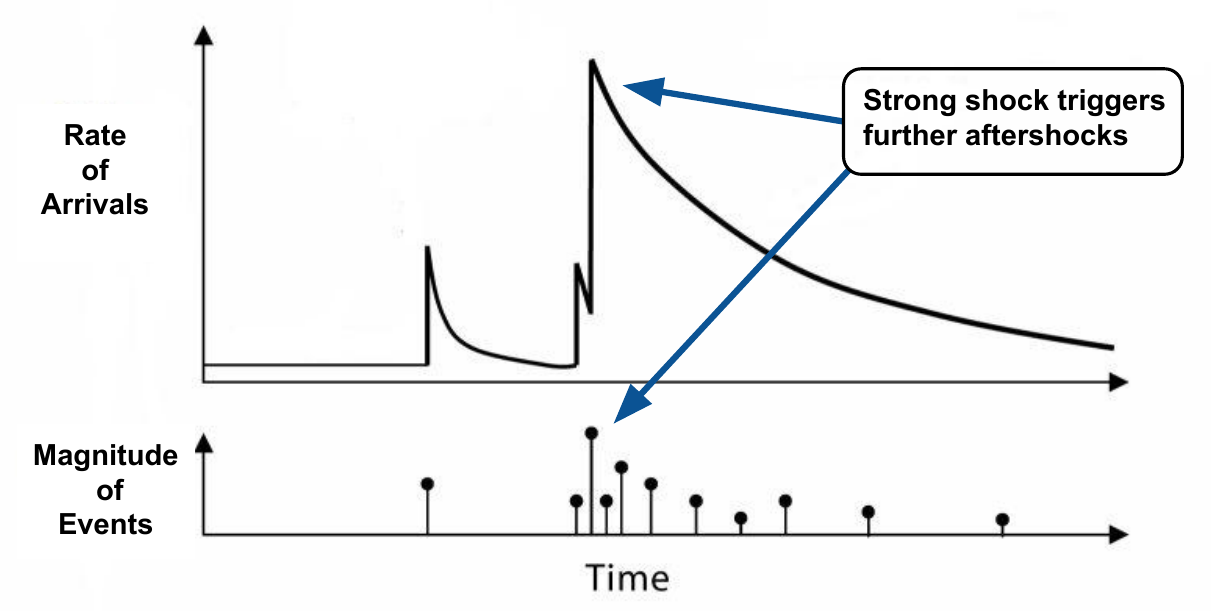}
\caption{Several phenomena, such as Earthquakes, exhibit temporal dependencies which can be modeled by Hawkes Processes. The diagram is modified from `\textit{quakecatcher.net}.' }\label{fig: qcatcher}
\end{figure}

\begin{figure*}[t!]
\centering
\includegraphics[width=0.97\linewidth]{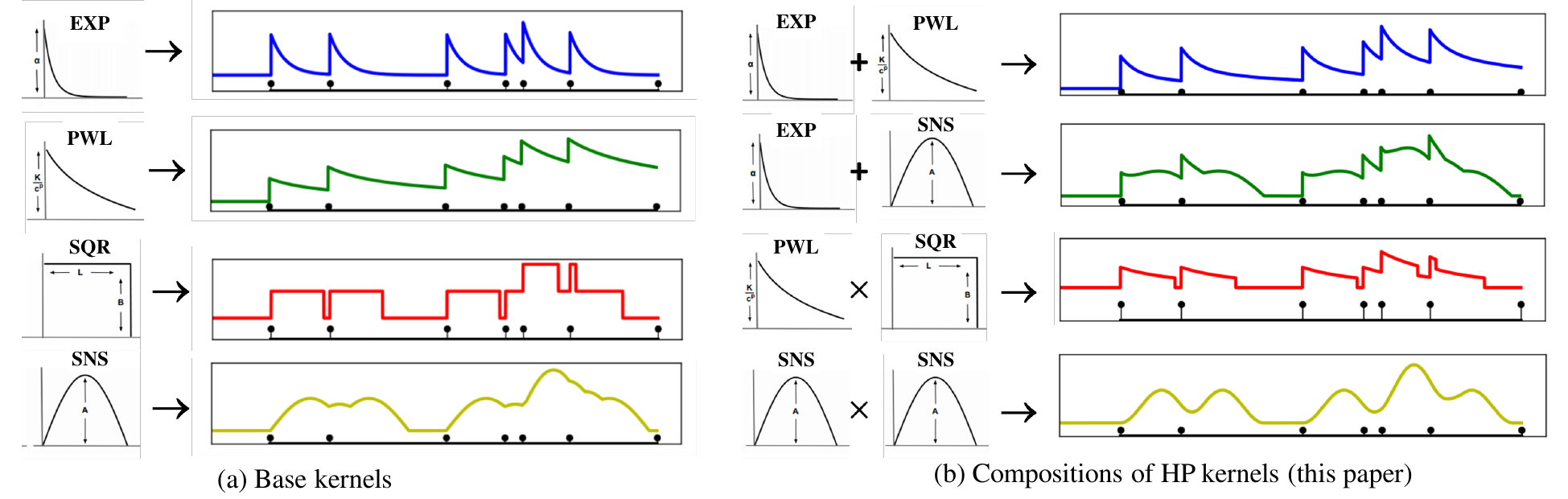}
\caption{\small Examples of intensity functions with four different self-triggering Point Processes $\lambda (t)$. The same observation could correspond to significantly different intensity functions, each one with distinct self-triggering kernels.}
\label{fig: lambda}
\end{figure*}

\section{Related Work}
Automatic analysis frameworks for Gaussian Processes (GPs) are proposed in \cite{DD13} and \cite{YH16}. However, due to fundamental distinctions between GPs and HPs (such as stationarity conditions and causality assumptions for the latter), the techniques proposed for GPs can not be extended to HPs in a straightforward manner. 

\cite{JE16} uses exponential kernels for modeling quick-decay in finance or web data. \cite{YO99} models slow decay influence with power-law kernels in earthquake, while \cite{QZ15} performs power law modeling experiments with social media-related data. \cite{HX16} uses sinusoidal kernels for modeling periodicity-inducing influence in TV watching-related data (IPTV) , in which watching one episode of a TV program makes the viewer more likely to watch further ones. Since these shows are usually broadcasted weekly, the TV-watching behavior is likely to demonstrate a weekly self-excitement. In addition, according to \cite{SL14}, homicide rates show a pronounced seasonal effect, peaking in the summer and tapering in the winter.

More recent works, such as the neural network-based  Hawkes processes in \cite{ND16,HM17} and the time-dependent Hawkes process (TiDeH) \cite{RK16}, allow for learning very flexible Hawkes processes with highly complicated intensity functions, while depending on the size and the quality of data. In this work, however, we focus on the interpretability, or explainability, of said functions and their corresponding typical behaviours, which are core factors on the Hawkes kernel selection and optimization.

\section{Hawkes Processes}

A point process with a sequence of n time-events is  expressed by a vector of the form ($t_{1}$,$t_{2}$, ... , $t_{n}$). Treating the real line as a time axis, the vector can be intuitively associated with a counting process $N(t)$, such that $dN(t) = 1$, if there is an event at time t; and $dN(t) = 0$, otherwise.

A point process can be described through its intensity function ($\lambda (t)$), which can be understood as the instantaneous expected rate of arrival of events, or the expectation of derivative of the counting process $N(t)$:
\begin{eqnarray}
\lambda (t) = \lim_{h \to 0} \frac{\mathbb{E} [N(t+h)-N(t)]}{h}
\end{eqnarray}

This intensity function uniquely characterizes the finite-dimensional distributions of the point process \cite{VJ03}. A simple example of this function would be the constant mean rate of arrival, $\mu$, in the case of a homogeneous Poisson process.

HPs model the intensity function in terms of \textit{self-excitation}: the arrival of an event makes subsequent arrivals more likely to happen \cite{PL15}. HPs can be described through the following conditional intensity function $\lambda (t)$:
\begin{eqnarray*}
\lim_{h \to 0} \dfrac{\mathbb{E} [N(t + h) - N(t)| \mathcal{H} (t)]}{h} \notag = \mu + \int_{-\infty}^{t} \phi (t-u) dN(u),
\end{eqnarray*}
where
\begin{itemize}
\item $\mathcal{H} (t)$ is the history of the process, the set containing all the events up to time t;
\item $\mu$ is called \textit{background rate}, or \textit{exogenous intensity}, which is fixed as the mean rate of a homogeneous Poisson process;
\item $\phi (t)$ is denominated \textit{self-triggering kernel}, or \textit{excitation function}.
\end{itemize}

From this function, one may notice that the intensity at time t will likely be affected by events which happened before the time t, described by the history of the process.
From \cite{AH71}, we have that, if:
\begin{eqnarray}
||\phi|| := \int_{0}^{\infty} \phi (t) dt \leq 1,
\end{eqnarray}
then the corresponding process will show wide-sense stationary behavior, from which the asymptotic steady arrival rate, or first-order statistics, $\Lambda = \tfrac{\mu}{(1-||\phi||)}$, 
can be obtained, along with its covariance function, or second-order statistics, which is independent of t, $\nu (\tau) =  \mathbb{E} [dN(t) dN(t + \tau)]$.






Estimating $\Lambda$ and $\nu (\tau)$ requires wide-sense stationarity assumptions which, besides being analytically convenient, are also connected to the fact that, in real data, the chain of self-excitedly induced further events will always be of finite type, or without `blowing up.' This corroborates the practicality of the estimated model.

\subsection{Discretized Kernel Estimation}

Being one possible way of recovering the triggering kernel of a HP, this step is fully described in \cite{BM12}, and basically consists of building an estimator of $\phi (t)$ from empirical measurements of $\nu (\tau)$, the stationary covariance.

Given a finite sequence of ordered time-events in $[0,T]$, we fix a window size of \textit{h}, and estimate $\nu (\tau)$ as:
\begin{eqnarray}
\nu_{\tau}^{(h)} = \dfrac{1}{h} E \left( (\int_{0}^{h} dN_{s} - \Lambda h) (\int_{\tau}^{\tau+h} dN_{s} -\Lambda h) \right)  
\end{eqnarray}

In practice, this estimation is done in discrete time steps $\delta$, up to a maximum value of $\tau$. \footnote{In our case, we used a carefully designed heuristics explained in the section Experimental Results.}:
\begin{eqnarray}
\nu_{\tau,\delta}^{(h)} {=} \dfrac{1}{T} \!\!\! \sum_{i=1}^{\lfloor T/\delta \rfloor} (dN_{i\delta}^{(h)}{-}dN_{(i-1)\delta}^{(h)}) (dN_{i\delta+\tau}^{(h)}{-}dN_{(i-1)\delta+\tau}^{(h)}),
\label{eq: grid}
\end{eqnarray}
where $dN_{i\delta}^{(h)}$ is the total number of events happening between $t=i \delta$ and $t=i \delta + h$.

From \cite{BM12}, we have that, given $g_{t}^{(h)} = (1-\frac{|h|}{t})^{+}$, i.e., a triangular kernel density estimator with bandwidth \textit{h}, we have the following relation in Laplace domain:
$\hat{\nu_{z}^{(h)}} = \hat{g_{z}}^{(h)} (1+\hat{\psi}_{z}^{\star}) \Lambda (1+\hat{\psi}_{z}^{\star})^{\dag},$ where\footnote{Given a function $f_{t}$, $\hat{f}_{z}$ is its Laplace Transform, and the ``$\star$'' symbol corresponds to its conjugate.}:
\begin{eqnarray*}
\hat{\psi}_{z} = \sum_{n=1}^{+\infty} \hat{\phi}_{z}^{n} = \tfrac{\hat{\phi}_{z}}{(1-\hat{\phi}_{z})}. 
\end{eqnarray*}
Working with the Fourier transform restriction, i.e., ($z=i \omega$, with $\omega \in \mathbb{R}$) and given that 
$\hat{g}_{i \omega}^{(h)} = \tfrac{4}{\omega^2 h} \sin^{2} (\tfrac{\omega h}{2}),$ we get to 
\begin{eqnarray}
 (1+\hat{\psi}_{z}^{\star}) \Lambda (1+\hat{\psi}_{z}^{\star})^{\dag} = \tfrac{\hat{\nu_{z}^{(h)}}}{\hat{g_{z}}^{(h)}},
\end{eqnarray}
where we fix $h = \delta$, so we do not bother with cancellations of $\hat{g}_{z}^{(h)}$.
Then, from: 
$|1+\hat{\psi}_{i \omega}|^2 {=} \tfrac{\hat{\nu}_{z}^{(h)}}{\Lambda \hat{g_{z}}^{(h)}},$ we get to the discretized estimation of $\phi_{t}$ by taking the inverse Fourier transform of:
\begin{eqnarray}
\hat{\phi}_{i \omega} = 1 - e^{-\log |1+\hat{\psi}_{i \omega}| + i H(\log |1+\hat{\psi}_{i \omega}|)},
\end{eqnarray}
in which the operator $H(\cdot)$ refers to the Hilbert transform.

\section{Automatic Kernel Decomposition for HPs}

This section presents the second step of the automatic kernel identification: a parametric kernel search through our new kernel decomposition scheme.

\subsection{Self-Exciting Kernels}

From the definition of the conditional intensity function, the self-excitation of the process is expressed through the kernel function $\phi(t)$. For the kernel decomposition, four base kernels will be used for identifying and estimating typical triggering behaviors as shown in Table \ref{tab: base_kernels}:
\begin{itemize}
\item \textbf{EXP($\alpha$,$\beta$)}: The decay exponential kernel is parameterized by the amplitude \textbf{$\alpha$} and decay rate \textbf{$\beta$}, and is useful for modeling quick influence decay, in which initial transactions/hyperlinks have a lot of impact initially but rapidly reduce their influence over time;
\item \textbf{PWL(K,c,p)}: The \textbf{p}o\textbf{w}er \textbf{l}aw kernel is parameterized by the amplitude \textbf{K}, the exponent \textbf{p}, and the constant \textbf{c}, modeling a slower decaying trend than the exponential;
\item \textbf{SQR(B,L)}: The  pulse kernel is described by the amplitude \textbf{B} and the length \textbf{L}. Being a trivial, steady, and self-exciting dynamics on its own, it may also work as an offset level for the combined triggering with other kernel types, in the case of addition, and as a horizon truncation, in the case of multiplication\footnote{$u(t)$ is the step function.};
\item \textbf{SNS($A$,$\omega$)}: A truncated \textbf{s}i\textbf{n}u\textbf{s}oidal kernel, parameterized by the amplitude \textbf{A} and the angular velocity \textbf{$\omega$}. This type of kernel base function captures well the self-excitement of periodic events.
\end{itemize}

\vspace{-2em}

\begin{figure}[H]
\centering
\includegraphics[width=\linewidth]{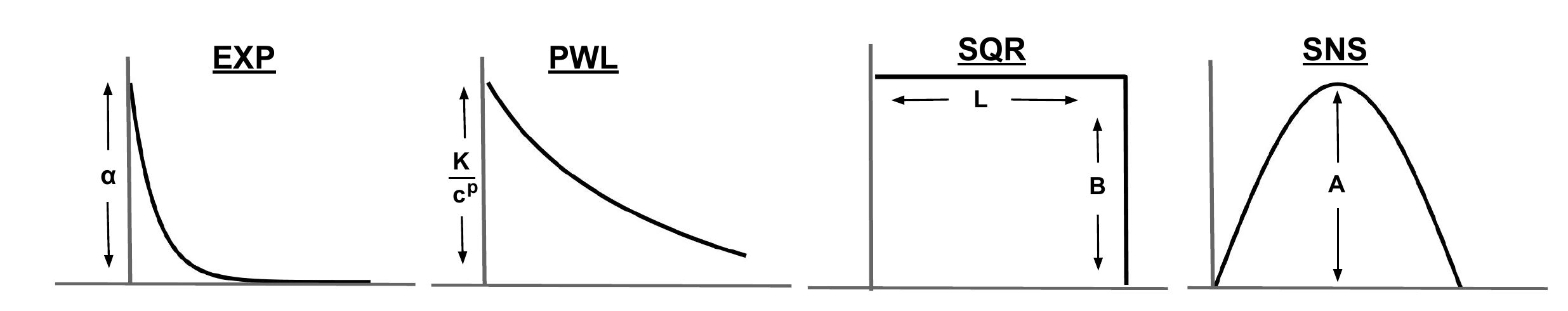}
\caption{Four base kernel types.}
\label{fig: Kernel_Plot}
\end{figure}

\begin{table}[H]
\centering
\renewcommand{\arraystretch}{1.3}
\begin{tabular}{l||p{3.3cm}}
\hline
\textbf{Type} & \textbf{Equation} \\
\hline
Exponential (EXP($\alpha$,$\beta$)) & 
$\alpha e^{-\beta t}$ \\
Power-Law (PWL(K,c,p)) & $\frac{K}{(c + t)^p}$, $(p>1)$ \\
Pulse (SQR(B,L)) & 
$B (u(t) - u(t-L))$ \\
Sinusoidal (SNS(A,$\omega$)) & $A sin(\omega t)$, $t \in \left[ 0,\frac{\pi}{\omega} 			\right]$\\
\hline
\end{tabular}
\caption{Base kernels and their equations.}\label{tab: base_kernels}
\end{table}

Here, the discretized kernel estimation is optional when a direct optimization of kernel structure is possible. Unfortunately, discontinuous functions (SQR, SNS) do not allow such optimization (e.g., Gradient Descent, Nelder-Mead). In this paper, we use the discretized kernel estimation as a unified method for both continuous (EXP, PWL) and discontinuous (SQR, SNS) kernels; and, most importantly, their combinations.

\begin{figure}[ht] 
  \begin{subfigure}[b]{0.33\linewidth}
    \centering
    \includegraphics[width=\linewidth]{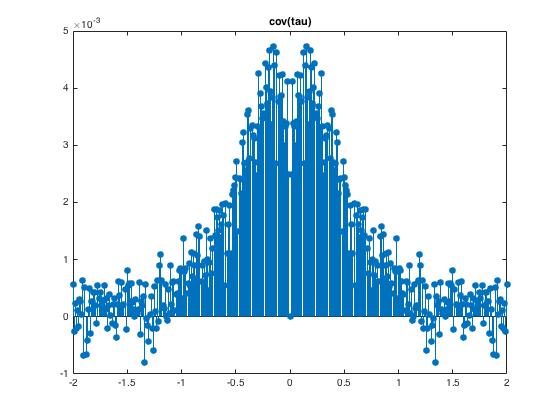} 
  \end{subfigure}
  \begin{subfigure}[b]{0.33\linewidth}
    \centering
    \includegraphics[width=\linewidth]{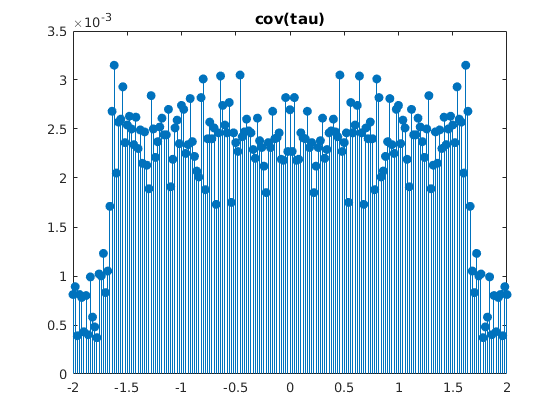}
  \end{subfigure} 
  \begin{subfigure}[b]{0.33\linewidth}
    \centering
    \includegraphics[width=\linewidth]{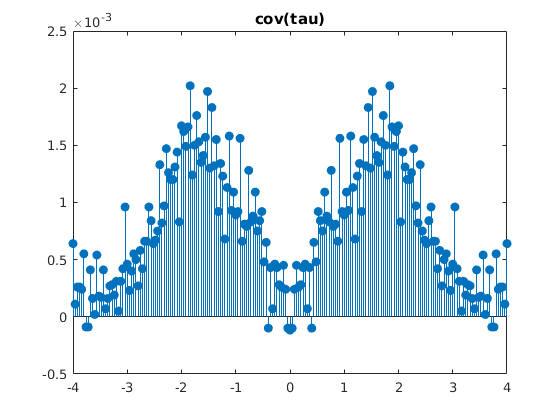}
  \end{subfigure}
  	\caption{Discretized covariance estimate from a sequence generated with 	(a) EXP, (b) SQR, and (c) SNS kernels.}
    \label{fig: covest}
\end{figure}

Furthermore, another great advantage of this step, compared with traditional sequential methods, is the fact that the value of $\nu$ for each value of $\tau$ can be calculated independently, while, in Gradient Descent, the value of the parameters at step t must be obtained before the values for step $t+1$. When combined with the parallelization of loops in our algorithm, this step significantly improves the speed of obtaining the most likely parametric representations of the sample processes.

\subsection{Kernel Decomposition}

\begin{figure}
\centering
\includegraphics[width=\linewidth]{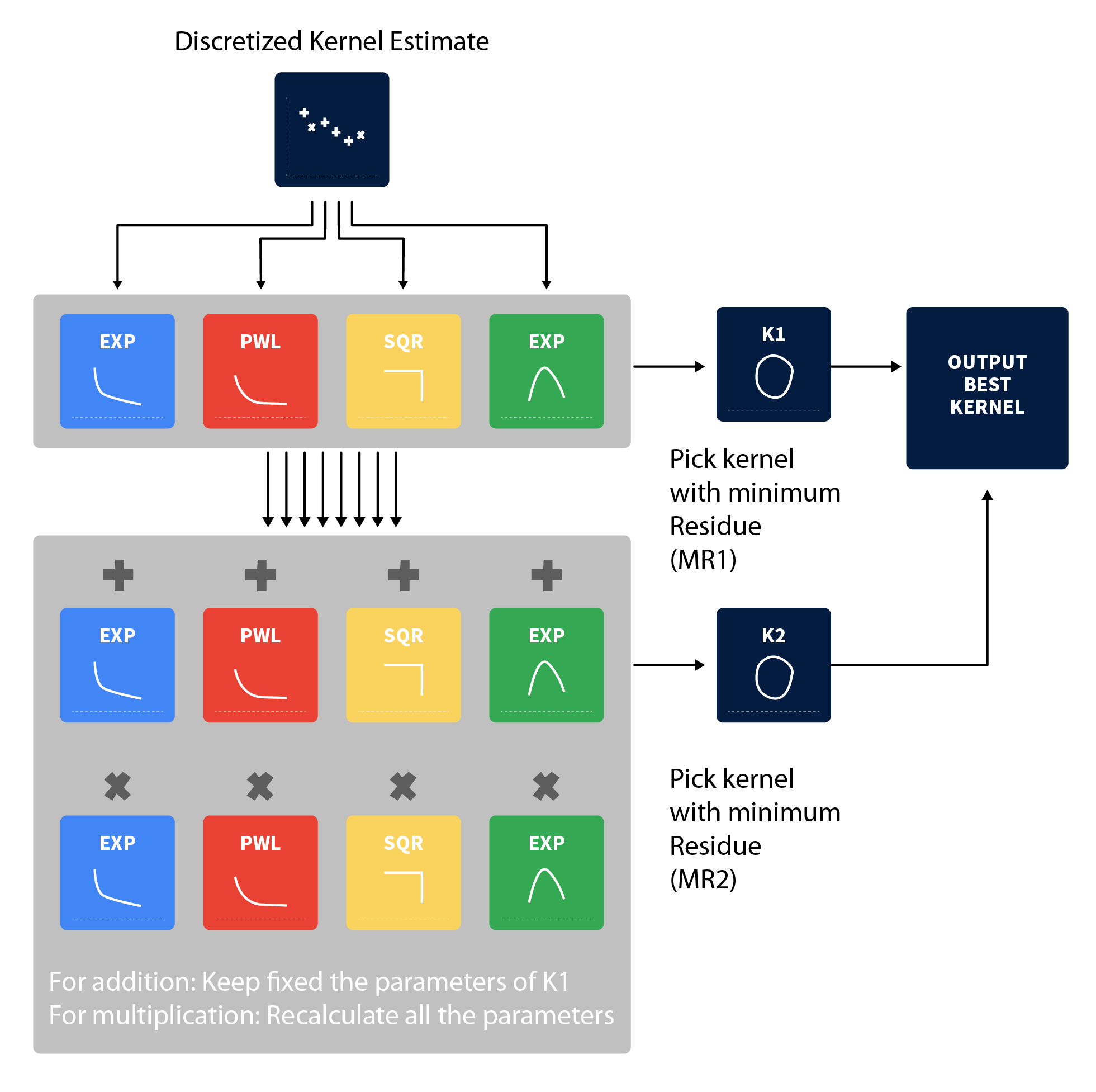}
\caption{Our kernel decomposition algorithm for exploiting and analyzing distinct behaviours in Hawkes Processes.}\label{fig: kernel_comp}
\end{figure}

For expressing the discretized estimation in terms of the four base kernels, the following steps are executed:
\begin{enumerate}
\item Calculate residues ($L^1$-error) w.r.t. the four basic kernels $\{$EXP, PWL, SQR, SNS$\}$;
\item Select the kernel with the minimum residue $MR_1$, denominated $K_1$;
\item Check whether the estimated parameters of the kernel satisfy the stationarity condition, by using the closed-form expressions from Table  \ref{tab: tab1};
\item Calculate residues w.r.t. a total of 8 kernel expansions, resulting from 2 operations (addition and multiplication) per base kernel $\{$+EXP, $\times$EXP, +PWL, $\times$PWL, +SQR, $\times$SQR, +SNS, $\times$SNS$\}$, while fixing the optimized parameters for $K_1$, in the case of Additive Combination, and recalculating all the parameters, in the case of Multiplicative Combination;
\item Select the kernel with minimum residue $MR_2$, denominated $K_2$, and check the spectral radius condition (calculated in closed-form from Table \ref{tab: tab2});
\item If both K1 and K2 are stable, and $MR_1 < MR_2/\eta$ ($\eta$ would act as a regularization parameter), pick $K_1$ . Else, pick $K_2$.
\item If likelihood (\textit{llh}) of direct optimization (GD, Nelder-Mead) is greater than likelihood of kernel decomposition, output GD model. Else, output the decomposition model.
\end{enumerate}
Regarding the computational efficiency of the decomposition algorithm, two strategies yielded results at a much lower computational cost, without altering the results of the decomposition:
\begin{itemize}
\item Selecting the best kernel through the error, instead of likelihood;
\item Greedy search of K2 based on the selected K1, instead of doing a brute-force search over all the $4 \times 8 = 32$ possible combinations for K2.
\end{itemize}
Figure \ref{fig: kernel_comp} explains the algorithm up to the depth two, for illustration purposes. Our kernel decomposition scheme can be expanded into multiple depths, as explained in Section \ref{sec:higher_order}. Our algorithm is presented in Algorithm \ref{alg:AKD}.
\begin{table}[H]
\centering
\renewcommand{\arraystretch}{1.0}
\begin{tabular}{c||c}
\hline
\textbf{Type} & \textbf{Stationarity Condition} \\
\hline
EXP($\alpha$,$\beta$) & 
$\alpha/\beta$\\
PWL(K,c,p) & $K c^{1-p}/(p-1$), $(p>1)$ \\
SQR(B,L) & 
$BL$ \\
SNS(A,$\omega$) & $2A/\omega$\\
\hline
\end{tabular}
\caption{Stationarity conditions of the  base kernels: the expressions in the stationarity condition should be set to less than 1.}\label{tab: tab1}
\end{table}
\begin{algorithm}[ht!]
\centering
\caption{Automatic Decomposition of HP Kernels}\label{alg:AKD}
\begin{algorithmic}[1]
\STATE $k_{est} \leftarrow input$, $output = Null$
\STATE $fit_1 \leftarrow \mbox{fit}(k_{est};\emptyset,\{$EXP, PWL, SQR, SNS$\})$
\STATE $K_1 \leftarrow \mbox{index\_of\_kernel}(\mbox{min\_residue}(fit_1))$
\STATE $MR_1 \leftarrow \mbox{min\_residue}(fit_1)$

\STATE $fit_2 \leftarrow \mbox{fit}(k_{est};K_1,\{\mbox{+EXP}, \mbox{* EXP}, \mbox{+PWL}, \mbox{*PWL},$ \par \hspace{36pt} $\mbox{+SQR}, \mbox{*SQR}, \mbox{+SNS}, \mbox{*SNS}\})$
\STATE $MR_2 \leftarrow \mbox{min\_residue}(fit_2)$
\STATE $K_2 \leftarrow \mbox{index}(\mbox{min\_residue}(fit_2))$
\IF{$||\phi_{K_1}|| < 1$}
	\STATE $output \leftarrow K_1$
\ENDIF

\IF{$||\phi_{K_2}|| < 1$}
	\IF{$output \neq Null$}
		\IF{$MR_1 \geq \frac{1}{\eta}MR_2$}
			\STATE $output \leftarrow K_2$
		\ENDIF
	\ELSE
		\STATE $output \leftarrow K_2$
	\ENDIF
\ENDIF

\IF{$\mbox{llh}(output) < \mbox{llh}(GD)$}
	\STATE $output \leftarrow GD$
\ENDIF

\end{algorithmic}
\end{algorithm}

\subsection{Stationarity Conditions}

\begin{table*}[t!]
\centering
\resizebox{1\textwidth}{!}{
\begin{tabular}{c| c||c}
\hline
\textbf{Base Kernel} & \textbf{Base Kernel} & \textbf{Condition}\\
\hline
\hline
 EXP($\alpha$,$\beta$) &  EXP($\alpha$,$\beta$) & $ \alpha_{1} \alpha_{2}/(\beta_{1}+\beta_{2})$  (closed under multiplication)\\
EXP($\alpha$,$\beta$) & PWL(K,c,p) & $\alpha K \beta^{p-1} e^{\beta c}\Gamma (1-p,\beta c) $\\
 EXP($\alpha$,$\beta$) &  SQR(B,L) &  $\left(\alpha B (1-e^{-\beta L})\right)/{\beta}$\\
EXP($\alpha$,$\beta$) & SNS(A,$\omega$) & $\left(A \alpha \omega (1 + e^{\frac{-\beta \pi}{\omega}})\right)/{(\omega^2 + \beta^2)}$\\
 PWL$(K_{1},c_{1},p_{1})$ &  PWL$(K_{2},c_{2},p_{2})$ &  $\leq \left( K_{1} K_{2}\right)/\left((p_{1}+p_{2}-1) \min(c_{1},c_{2})^{(p_{1}+p_{2}-1)}\right)$ (upper bound)\\
PWL(K,c,p) & SQR(B,L) & $\left(KB (c^{-(p-1)} - (c+L)^{-(p-1)})\right)/(p-1)$ \\
 PWL(K,c,p) &  SNS(A,$\omega$)  &  $\leq KA\left((c + \frac{\pi}{\omega})^{1-p} - c^{1-p}\right)/(1-p)$ (upper bound)\\
SQR(B,L) & SQR(B,L) & $BL$\\
 SQR(B,L) &  SNS(A,$\omega$) &  $2AB/\omega$\\
SNS(A,$\omega$) & SNS(A,$\omega$) & $\pi A/(2 \omega)$\\\hline
\end{tabular}%
 }
\caption{Stationarity Condition for Multiplicative Combination of the four Base Kernels.} \label{tab: tab2}
\end{table*}
Verifying the stationarity condition is one of the most important steps in the kernel search. When we end up with a non-stationary kernel, estimating future events can not be accurate.

To solve this issue, we developed closed-form expressions, either in the form of equality or as an upper bound, which are shown in Table \ref{tab: tab1}, for the case of a single kernel, and Table \ref{tab: tab2}, for multiplicative combinations of two kernels \footnote{ $\Gamma(\cdot,\cdot)$ is the well-known incomplete Gamma function: $\Gamma (a,y) = \int_{y}^{\infty} t^{a-1} e^{-t} dt$}. The conditions for additive combinations can be derived from the conditions for single kernels in a straightforward manner.

The kernel is said to induce stationarity if the result of the expression calculated using the estimated parameters belongs to the interval [0,1). This can be justified both from the point-of-view of HP as a branching process, also called \textit{immigrant-birth representation} \cite{TL09}, and of the boundedness of the spectral radius (largest absolute value among the eigenvalues) of the excitation matrix. \footnote{For the univariate HP case, the excitation matrix has dimension one, being only the excitation function, $\phi(t)$.}

\subsection{Scale-Independence Criterion}
\label{sec:scale-independece}

For an automatic time series analysis, scale-independence is indispensable, as time sequences of disjoint datasets may occur in time scales differing by several orders of magnitude. As an example, earthquake events' occurrences in a sequence are spaced by intervals of monthly and yearly scales. Thus, setting a horizon of a few months as the maximum value of $\tau$ in Equation (\ref{eq: grid}) might result in a satisfactory discrete estimation grid. However, using the same time length for estimating the triggering behavior of a finance-related sequence would require an impractically large grid resolution.

A histogram of all the time intervals between events in a sequence may be readily generated, and is an indicator of the overall magnitude of the spacing among the events. Thus, as a rule of thumb, the horizon length for $\tau$ may be set as the smaller time interval strictly larger than a percentage of the sequence's intervals. The values of 50 \% and 95 \% were used. In practice, this value of horizon length is obtained with the help of a histogram composed by 100 bins.

\section{Higher-order Kernel Decomposition}\label{sec:higher_order}

A sequential additive decomposition of the discretized estimation vector is rather straightforward, since one may just set the residual vector from the previous stages as the input of the next ones. 

In the case of multiplicative decomposition, it is nontrivial to find the result of intraclass decomposition. To the best of our knowledge, no analysis on multiplicative HP kernel decomposition is reported yet. 

In this paper, we provide a new upper bound over an  interclass kernel product of unknown degree, as in:
$$\left[\mbox{EXP}\right]^{k_{1}} {\times} \left[\mbox{PWL}\right]^{k_{2}} {\times} \left[\mbox{SQR}\right]^{k_{3}} {\times} \left[\mbox{SNS}\right]^{k_{4}}$$
for $k_i \in \mathbb{Z}^{*}$,
where the operator ``$\left[ \cdot \right]^{k}$'' corresponds to the set of functions which can be decomposed into a k-th order product of kernels, e.g:
$${\left[EXP\right]}^{k} = \underbrace{\alpha_1 e^{-\beta_1 x} * \alpha_2 e^{-\beta_2 x} * ... * \alpha_{k} e^{-\beta_{k} x}}_{\text{k terms}}.$$
By deriving the four possible intraclass kernel products, one may observe that the typical self-exciting behavior features of each kernel type are preserved, as in the following:
\begin{itemize}
\item $\left[\mbox{EXP}\right]^{k_{1}}$ reduces to the case of a single exponential with $\alpha = \prod_{i=1}^{k_{1}} \alpha_i$ and $\beta = \sum_{i=1}^{k_{1}} \beta_i$, thus still accounting for its `quick-decay' behavior: $\left[\mbox{EXP}\right]^{k_1} \subset \left[\mbox{EXP}\right]$
\item $\left[\mbox{PWL}\right]^{k_{2}}$ is lower bounded by a single PWL kernel with $K = \prod_{i=1}^{k_{2}} K_i$, $c = max(c_1,...,c_{k_{2})}$ and $p = \sum_{i=1}^{k_2} p_i$, thus still accounting for its `slow-decay' behavior
\item $\left[\mbox{SQR}\right]^{k_{3}}$ reduces to a single SQR kernel with $B = \prod_{i=1}^{k_{4}} B_{i}$ and $L = min(L_1,...,L_{k_{4}})$, thus still accounting for its `steady-triggering' behavior: $\left[\mbox{SQR}\right]^{k_3} \subset \left[\mbox{SQR}\right]$
\item $\left[\mbox{SNS}\right]^{k_{4}}$ has $A = \prod_{i=1}^{k_{4}} A_i$ and a `spikier' aspect (higher bandwidth), thus still accounting for its `periodicity-inducing' behavior
\end{itemize}
Thus, on deepening the decomposition algorithm by overly increasing the number of levels above 2, we may be, in fact, adding little information on the qualitative aspect of the self-exciting behavior analysis of the data while making it more prone to overfitting to the noisiness of the discretized estimation vectors.

\subsection{Upper Bound}
Furthermore, regarding the boundedness of the higher-order decompositions, from the exact results for EXP and SQR intraclass decompositions and the upper bounds for the PWL and SNS ones, we have that:
\begin{eqnarray}
&&\left[\mbox{EXP}\right]^{k_{1}}{\times} \left[\mbox{PWL}\right]^{k_{2}}{\times} \left[\mbox{SQR}\right]^{k_{3}}{\times} \left[\mbox{SNS}\right]^{k_{4}} \notag \\
&\leq& \alpha e^{-\beta x} \dfrac{K}{(x+c_{upper})^p} B A sin(\omega x) 
\notag \leq \dfrac{\alpha B K A e^{-\beta x}}{(x+c_{upper})^p} \notag \\
&=& \mbox{EXP}(\alpha,\beta) {\times} \mbox{PWL}(K,c_{upper},p)^{k_2}{\times} \mbox{SQR}(B,L) {\times} A \notag,
\end{eqnarray}
for $0 \leq x \leq min(L,\dfrac{\pi}{\omega})$, and 0 otherwise.

\section{Experimental Results}\label{sec:Exp}




To demonstrate the benefits of the kernel decomposition framework, we conducted experiments with synthetic, financial and earthquake data.

For real-world data sets, no prior information about the kernel (type and parameters) is available. Thus, we use the log-likelihood of the kernel function over the time sequence as a quality criterion.

Given a realization $(t_1,t_2,...,t_k)$ of some regular point process on [0,T], its log-likelihood (\textit{l}) is expressed as: $l(t_1,...,t_k) = \sum_{i=1}^{k} \log (\lambda (t_i)) - \int_{0}^{T} \lambda (u) du$.

For an automatic time series analysis, scale-independence is indispensable, as time sequences of disjoint datasets may occur in time scales differing by several orders of magnitude. As an example, earthquake events' occurrences in a sequence are spaced by intervals of monthly and yearly scales. Thus, setting a horizon of a few months as the maximum value of $\tau$ in Equation (\ref{eq: grid}) might result in a satisfactory discrete estimation grid. However, using the same time length for estimating the triggering behavior of a finance-related sequence would require an impractically large grid resolution.

A histogram of all the time intervals between events in a sequence may be readily generated, and is an indicator of the overall magnitude of the spacing among the events. Thus, as a rule of thumb, the horizon length for $\tau$ may be set as the smaller time interval strictly larger than a percentage of the sequence's intervals. The values of 50 \% and 95 \% were used. In practice, this value of horizon length is obtained with the help of a histogram composed by 100 bins.


\subsection{Financial Data}

In the finance domain, HPs have become more prevalent, due to its structure being naturally adapted to model systems in which the discrete nature of the jumps in $N_{t}$ is relevant, making the model remarkably suited to modeling high-frequency data \cite{EB15}.

Here, we picked the 19 top-varying companies of the Technology, Healthcare, Industrial, Services and Utilities categories of Yahoo Finance. We extracted tick data from every two minutes of 30 business days (02/02/2017 to 02/23/2017 for Technology and 04/07/2017 to 05/18/2017 for the other ones). Whenever a stock price changed by some magnitude higher than some threshold, an event was logged in the corresponding time sequence. Ten different percentual thresholds, increasing at equally spaced intervals from 0.03\% to 0.3\%, were applied. This procedure resulted in a number of valid sequences, for each category, indicated in Table \ref{tab: stockexp}, since the remaining ones did not contain enough points for the splitting between training and validation subsequences.

As in extrapolation tasks, the 80\% of the first elements for each sequence were then used as training data, and the remaining 20\% were used for validation, i.e., we estimated the parameters of the kernel using the first 24 days and then calculated the log-likelihood on the last 6 days of each sequence. The kernel was then normalized to 2 min = 120 sec. When comparing the log-likelihoods of first and second level decompositions, we observed that the second level, with composite kernels, resulted in a higher log-likelihood in the majority of sequences from each category, as indicated in Table \ref{tab: stockexp}, what corroborates that a more flexible model of the kernel provides a more accurate description of the underlying dynamics of the process. The average log-likelihood for each level is shown in Table \ref{tab: comp_K1vsK2_financial}. The comparison of each sequence is shown in Figure \ref{fig: comp_K1K2_llh}.

When comparing the performance of the best estimation among the two levels and the usual exponential HP model used in financial analysis, fitted through the gradient-based method from \cite{TO79}, it is possible to see that the kernel composition exhibited a much more robust performance. Although the exponential HP performed well in some sequences, it tended to get stuck in local maxima with very poor performance, usually leading to unstable or negative combinations of parameters, for which the likelihood is null. The kernel composition performs better in the majority of sequences, as shown in Table \ref{tab: stockexp}. We provide the comparison of each individual sequence in the supplementary material. 
\begin{table}
\centering
\arrayrulecolor{black}
\begin{tabular}{ c || c  c }
\hline
Dataset & l(K1) & l(K2) \\
\hline
Technology & -2097.0 & -1894.6 \\ 
Healthcare & -2677.7 & -2446.2 \\ 
Industrial & -2309.7 & -2127.9 \\ 
Services & -2368.9 & -2218.7 \\ 
Utilities & -2427.9 & -2266.3 \\ 
\hline
\end{tabular}
  \caption{Average log-likelihood, over each of the five financial data sets, for the two levels of decomposition.}
  \label{tab: comp_K1vsK2_financial}
\end{table}
\begin{figure}[h!]
\centering
\includegraphics[width=\linewidth]{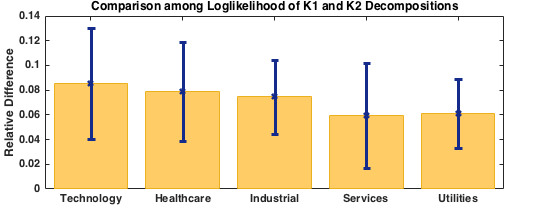} 
\caption{Error bar plot with mean and standard deviation of the difference among log-likelihood, $l(K2) - l(K1)$, over each of the five categories of stocks. A positive value means that l(K2) is greater than l(K1).}
\label{fig: comp_K1K2_llh} 
\end{figure}
\begin{table}
\centering
\setlength{\tabcolsep}{0.2em}
\begin{tabular}{c|c|c|c}
\hline
Dataset & \# of Seq. & \shortstack{$l(\mbox{K1}{,}\mbox{K2}){>}l(\text{EXP})$} & \shortstack{$l(\mbox{K2}) {>} l(\mbox{K1})$}\\
\hline
Technology & 70 &  67.14\% & 98.57\%\\
Healthcare & 117 & 62.39\% & 92.31\%\\
Industrial & 53 & 64.15\% & 94.34\%\\
Services & 61 & 54.09\% & 85.25\%\\
Utilities & 48 & 77.08\% & 93.75\% \\
\hline
\end{tabular}
\caption{Aggregate comparison, among the gradient descent based HP model and the first- and second-level decompositions of the proposed algorithm, for each of the five financial data sets.}\label{tab: stockexp}
\end{table}
\begin{figure}[ht!]
\centering
\includegraphics[width=\linewidth,height=1.7cm]{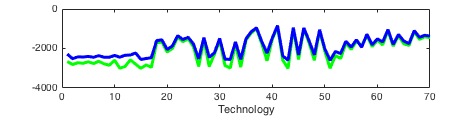} 
\includegraphics[width=\linewidth,height=1.7cm]{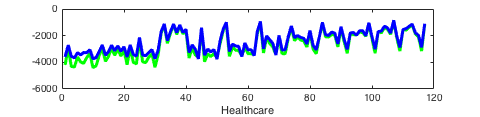} 
\includegraphics[width=\linewidth,height=1.7cm]{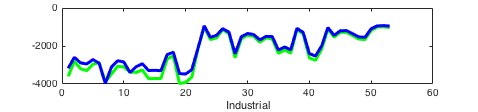} 
\includegraphics[width=\linewidth,height=1.7cm]{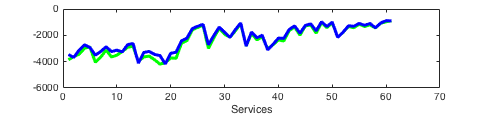} 
\includegraphics[width=\linewidth,height=1.7cm]{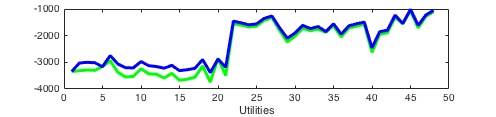} 
\caption{Comparison among log-likelihood of first ({\color{green} green}) and second level ({\color{blue} blue}) of kernel decomposition algorithm, for each valid sequence.}
\label{fig: comp_K1K2_llh} 
\end{figure}

\subsection{Earthquake Data}

\begin{figure*}[ht!]
\centering
\begin{subfigure}[b]{0.33\linewidth}
\centering
\includegraphics[width=\linewidth]{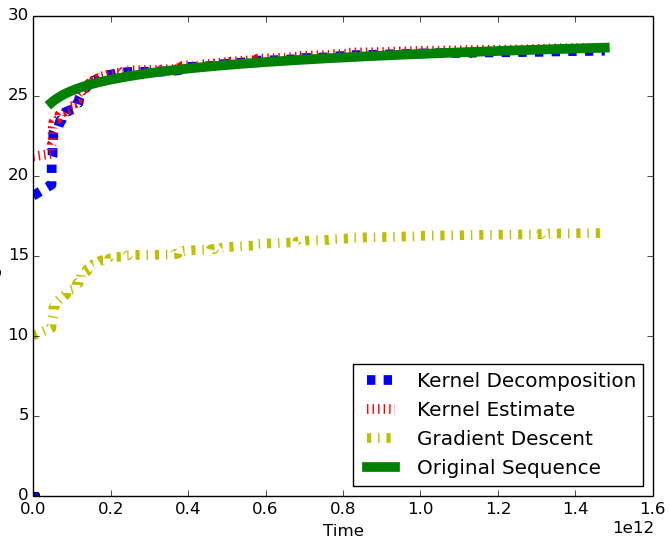} 
\end{subfigure}
    \hfill
\begin{subfigure}[b]{0.33\linewidth}
\centering
\includegraphics[width=\linewidth]{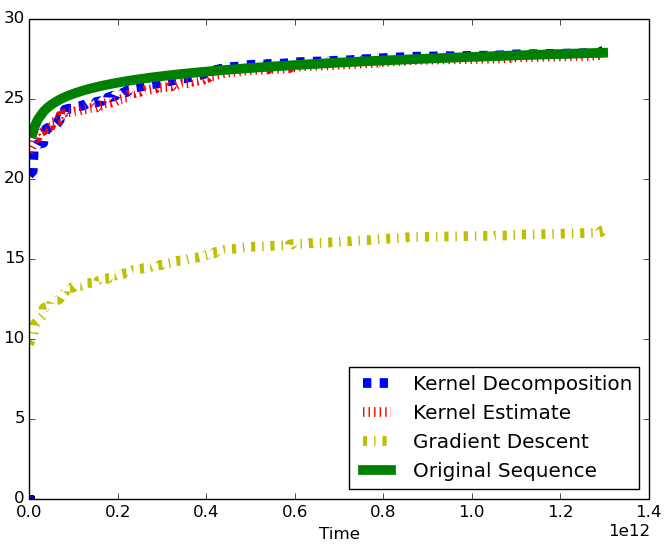} 
\end{subfigure}
\hfill
\begin{subfigure}[b]{0.33\linewidth}
\centering
\includegraphics[width=\linewidth]{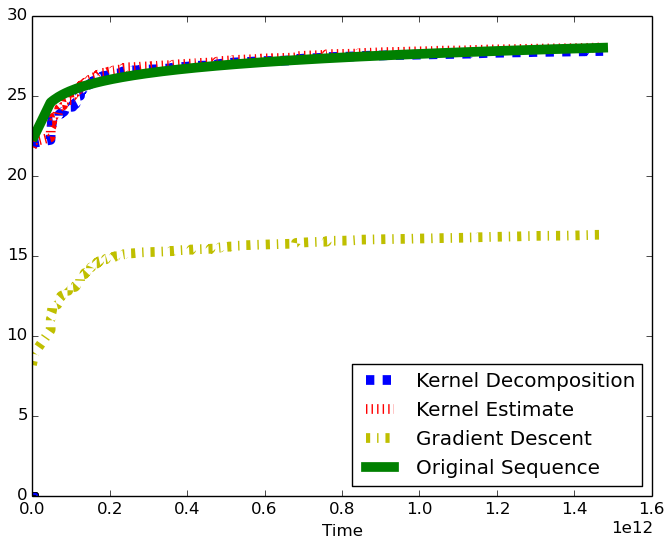} 
\end{subfigure}
\caption{Q-Q Plots, in logarithmic form (x axis is time and y axis is magnitude), from kernel decomposition estimation over the Earthquake Dataset sequences.}
\label{fig: qq_earthquake} 
\end{figure*}

The data considered for the earthquake experiment was a set of 100 time sequences extracted from the USGS NCSN Catalog (NCEDC database), from the day of 01/01/1966 to 01/01/2015. The latitude range was [30,55], and the longitude range was [-140,-110]. Different length intervals and resulting areas were considered. Whenever the magnitude of an event exceeded some threshold, its time coordinate was added to the corresponding input time sequence. The magnitude thresholds were varied among 2.5, 3.0, 3.5 and 4.0; and the grid resolution was set to 20 and 100 points.

Seeking scale-independent search, we use the aforementioned histogram heuristics: earthquakes events are separated by time intervals of monthly or yearly scales. Thus, the estimation horizon for financial data, for example, lasting usually only a few seconds, would hardly capture the overall aspect of the triggering behavior in this case.
\arrayrulecolor{black}

\begin{figure}[H] 
    \centering
  \begin{subfigure}[b]{0.85\linewidth}
    \centering
    \includegraphics[width=\linewidth]{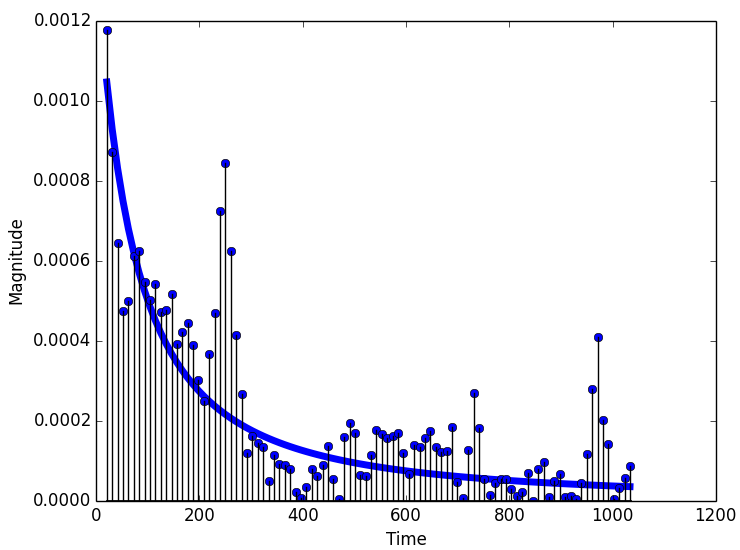}
    \caption{Stocks}
  \end{subfigure}
  
  \begin{subfigure}[b]{0.8\linewidth}
    \centering
    \includegraphics[width=\linewidth]{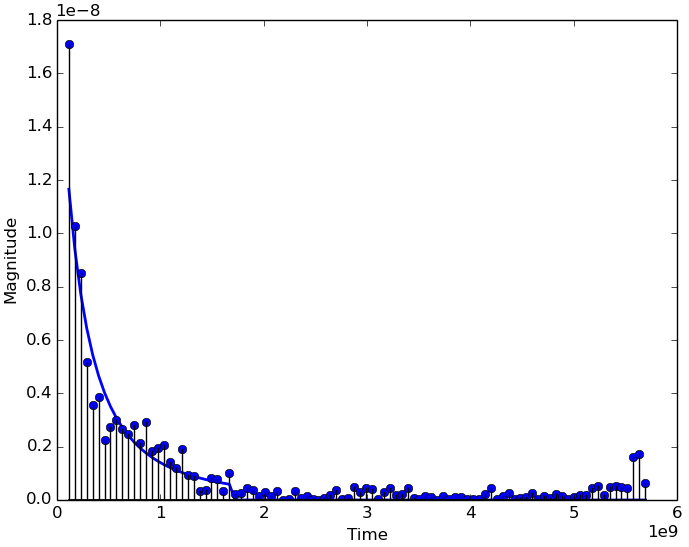}
    \caption{Earthquake}
  \end{subfigure} 
  \caption{Good resolution of kernels in (a) Stocks and (b) Earthquake data was achieved by histogram-based criteria, despite the very different time scales among these two kinds of data ($10^3$ and $10^9$, respectively).}
\label{fig: hist_trick} 
\end{figure}
The results indicate a strong agreement with the long standing assumption of a power-law shaped kernel for the intensity of aftershocks' occurrences (`Omori's Law' (1894)). For 20-point grid resolution, the relative frequency of each kernel was $\mbox{(EXP, PWL, SQR, SNS)}=\mbox{(0, 97, 2, 1)}$. For the 100-point grid resolution, the relative frequency was \mbox{(0, 99, 1, 0)}. Q-Q plots from the estimated models are shown in Figure \ref{fig: qq_earthquake}, in which comparisons to the original sequence are made among sequences generated by our kernel composition, disconsidering the stabiliy check, the discretized estimate and the usual Power-Law kernel model, fitted through the gradient descent based method (GD) \cite{YO99}. Both our method and the discretized estimate perform very close to the original sequence, while the GD method tended to get stuck in local optima with poor performance. A timescale-based initialization of $\mu$ was used.









\subsection{Verifying Scale-Independence Criterion}
To verify the histogram criteria (explained in Section~\ref{sec:scale-independece}), used the data sets ( Stocks and Earthquake). As shown in Figure~\ref{fig: hist_trick}, The histogram criteria allows us to find a good resolution of kernels in highly different scales.

\section{Conclusion}
Hawkes processes are point processes which capture self-exciting discrete events in time series data. To predict future events with HPs, an appropriate kernel is selected by hands, previously. In this paper, we proposed a new temporal covariance-based kernel decomposition method to represent various self-exciting behaviors. We also presented a model (structure/parameter) learning algorithm to select the best HP kernel given the temporal discrete events. The stationarity conditions are derived to guarantee the validity of the kernel learning algorithm. In experiments, we demonstrated that the proposed algorithm performs better than existing methods to predict future events by automatically selecting kernels.
\clearpage
{
\bibliography{abcd_HP}

\begin{thebibliography}{}

\bibitem[\protect\citeauthoryear{Bacry and Muzy}{2016}]{BM16}
Bacry, E., and Muzy, J.
\newblock 2016.
\newblock First- and second-order statistics characterization of hawkes
  processes and non-parametric estimation.
\newblock {\em {IEEE} Transactions on Information Theory} 62(4):2184--2202.

\bibitem[\protect\citeauthoryear{Bacry, Dayri, and Muzy}{2012}]{BM12}
Bacry, E.; Dayri, K.; and Muzy, J.~F.
\newblock 2012.
\newblock Non-parametric kernel estimation for symmetric hawkes processes.
  application to high frequency financial data.
\newblock {\em The European Physical Journal B} 85(5):1--12.

\bibitem[\protect\citeauthoryear{Bacry, Mastromatteo, and Muzy}{2015}]{EB15}
Bacry, E.; Mastromatteo, I.; and Muzy, J.-F.
\newblock 2015.
\newblock Hawkes processes in finance.
\newblock {\em Market Microstructure and Liquidity} 01(01):1550005.

\bibitem[\protect\citeauthoryear{Daley and Vere-Jones}{2003}]{VJ03}
Daley, D., and Vere-Jones, D.
\newblock 2003.
\newblock {\em An Introduction to the Theory of Point Processes: Volume I:
  Elementary Theory and Methods}.
\newblock Springer.

\bibitem[\protect\citeauthoryear{Du \bgroup et al\mbox.\egroup }{2016}]{ND16}
Du, N.; Dai, H.; Trivedi, R.; Upadhyay, U.; Gomez{-}Rodriguez, M.; and Song, L.
\newblock 2016.
\newblock Recurrent marked temporal point processes: Embedding event history to
  vector.
\newblock In {\em Proceedings of the {ACM} {SIGKDD} International Conference on
  Knowledge Discovery and Data Mining},  1555--1564.

\bibitem[\protect\citeauthoryear{Duvenaud \bgroup et al\mbox.\egroup
  }{2013}]{DD13}
Duvenaud, D.~K.; Lloyd, J.~R.; Grosse, R.~B.; Tenenbaum, J.~B.; and Ghahramani,
  Z.
\newblock 2013.
\newblock Structure discovery in nonparametric regression through compositional
  kernel search.
\newblock In {\em Proceedings of the International Conference on Machine
  Learning},  1166--1174.

\bibitem[\protect\citeauthoryear{Embrechts, Liniger, and Lin}{2011}]{TL11}
Embrechts, P.; Liniger, T.; and Lin, L.
\newblock 2011.
\newblock Multivariate hawkes processes: an application to financial data.
\newblock {\em Applied Probability Trust}.

\bibitem[\protect\citeauthoryear{Etesami \bgroup et al\mbox.\egroup
  }{2016}]{JE16}
Etesami, J.; Kiyavash, N.; Zhang, K.; and Singhal, K.
\newblock 2016.
\newblock Learning network of multivariate hawkes processes: {A} time series
  approach.
\newblock In {\em Proceedings of the Conference on Uncertainty in Artificial
  Intelligence},  162--171.

\bibitem[\protect\citeauthoryear{Hawkes}{1971}]{AH71}
Hawkes, A.~G.
\newblock 1971.
\newblock Spectra of some self-exciting and mutually exciting point processes.
\newblock {\em Biometrika} 58(1):201--213.

\bibitem[\protect\citeauthoryear{Hwang, Tong, and Choi}{2016}]{YH16}
Hwang, Y.; Tong, A.; and Choi, J.
\newblock 2016.
\newblock Automatic construction of nonparametric relational regression models
  for multiple time series.
\newblock In {\em Proceedings of the International Conference on Machine
  Learning},  3030--3039.

\bibitem[\protect\citeauthoryear{Kobayashi and Lambiotte}{2016}]{RK16}
Kobayashi, R., and Lambiotte, R.
\newblock 2016.
\newblock Tideh: Time-dependent hawkes process for predicting retweet dynamics.
\newblock In {\em Proceedings of the Tenth International Conference on Web and
  Social Media},  191--200.

\bibitem[\protect\citeauthoryear{Laub, Taimre, and Pollett}{2015}]{PL15}
Laub, P.; Taimre, T.; and Pollett, P.
\newblock 2015.
\newblock Hawkes processes.
\newblock {\em ArXiv e-prints}  1507.02822.

\bibitem[\protect\citeauthoryear{Linderman and Adams}{2014}]{SL14}
Linderman, S.~W., and Adams, R.~P.
\newblock 2014.
\newblock Discovering latent network structure in point process data.
\newblock In {\em Proceedings of the International Conference on Machine
  Learning},  1413--1421.

\bibitem[\protect\citeauthoryear{Liniger}{2009}]{TL09}
Liniger, T.
\newblock 2009.
\newblock {\em Multivariate Hawkes Processes}.
\newblock Ph.D. Dissertation, ETH Zurich.

\bibitem[\protect\citeauthoryear{Mei and Eisner}{2017}]{HM17}
Mei, H., and Eisner, J.
\newblock 2017.
\newblock The neural hawkes process: {A} neurally self-modulating multivariate
  point process.
\newblock In {\em Proceedings of the Annual Conference on Neural Information
  Processing Systems},  6754--6764.

\bibitem[\protect\citeauthoryear{Mohler \bgroup et al\mbox.\egroup
  }{2012}]{GM12}
Mohler, G.~O.; Short, M.~B.; Brantingham, P.~J.; Schoenberg, F.~P.; and Tita,
  G.~E.
\newblock 2012.
\newblock Self-exciting point process modelling of crime.
\newblock {\em Journal of the American Statistical Association}
  106(493):100--108.

\bibitem[\protect\citeauthoryear{Ogata}{1999}]{YO99}
Ogata, Y.
\newblock 1999.
\newblock Seismicity analysis through point-process modelling: A review.
\newblock {\em Pure and Applied Geophysics} 155(5):471--507.

\bibitem[\protect\citeauthoryear{Ozaki}{1979}]{TO79}
Ozaki, T.
\newblock 1979.
\newblock Maximum likelihood estimation of hawkes' self-exciting point
  processes.
\newblock {\em Annals of the Institute of Statistical Mathematics}
  (31):145--155.

\bibitem[\protect\citeauthoryear{Xu, Farajtabar, and Zha}{2016}]{HX16}
Xu, H.; Farajtabar, M.; and Zha, H.
\newblock 2016.
\newblock Learning granger causality for hawkes processes.
\newblock In {\em Proceedings of the International Conference on Machine
  Learning},  1717--1726.

\bibitem[\protect\citeauthoryear{Zhao \bgroup et al\mbox.\egroup }{2015}]{QZ15}
Zhao, Q.; Erdogdu, M.~A.; He, H.~Y.; Rajaraman, A.; and Leskovec, J.
\newblock 2015.
\newblock Seismic: A self-exciting point process model for predicting tweet
  popularity.
\newblock In {\em Proceedings of the ACM SIGKDD International Conference on
  Knowledge Discovery and Data Mining},  1513--1522.

\end{thebibliography}
\bibliographystyle{aaai}
}
\onecolumn
\makeatother
\vbox{
\centering
}
\appendix




\section{Derivations of Stationarity Criteria for Multiplicative Combinations of Kernels}

This appendix introduces the full derivations of stationarity criteria for the second order multiplicative compositions of the four base kernels.

\subsection{EXP x EXP}

For the combination ``EXPxEXP'', we have that, for stationarity to be achieved:
\begin{eqnarray}
0 \leq \int_{0}^{\infty} EXP(\alpha_1,\beta_1) EXP(\alpha_2,\beta_2) dx < 1 \nonumber
\end{eqnarray}
\begin{eqnarray}
0 \leq \int_{0}^{\infty} \alpha_1 e^{\alpha_1 x} \alpha_2 e^{\beta_2 x} dx < 1 \nonumber
\end{eqnarray}

Thus:
\begin{eqnarray}
\int_{0}^{\infty} \alpha_1 e^{-\beta_1 x} \alpha_2 e^{-\beta_2 x} dx &=& \int_{0}^{\infty} (\alpha_1 \alpha_2) e^{-(\beta_1 + \beta_2)x} dx \nonumber \\&=& 
\int_{0}^{\infty} \alpha e^{-\beta x} dx = \dfrac{\alpha}{\beta} = \dfrac{\alpha_1 \alpha_2}{\beta_1 + \beta_2} \nonumber
\end{eqnarray}

So, this case reduces to the case of a single exponential.

\subsection{EXP x PWL}

For the combination ``EXPxPWL'', we have that, for stationarity to be achieved:
\begin{eqnarray}
0 \leq \int_{0}^{\infty} EXP(\alpha,\beta) PWL(K,c,p) dx < 1 \nonumber
\end{eqnarray}
\begin{eqnarray}
0 \leq \int_{0}^{\infty} \alpha e^{-\beta x} \dfrac{K}{(x+c)^p} dx < 1 \nonumber
\end{eqnarray}
Thus:
\begin{eqnarray}
\int_{0}^{\infty} \alpha e^{-\beta x} \dfrac{K}{(x+c)^p} dx &=& \alpha K \int_{0}^{\infty} (x+c)^{-p} e^{-\beta x} dx \nonumber \\&=& \alpha K e^{\beta c} \int_{0}^{\infty} (x+c)^{-p} e^{-\beta (x+c)} dx \nonumber \\&=& \alpha K e^{\beta c} \beta^{p} \int_{0}^{\infty}(\beta (x+c))^{-p} e^{-\beta (x+c)} dx \nonumber \\&=& \alpha K e^{\beta c} \beta^{p-1} \int_{\beta c}^{\infty} t^{-p} e^{-t} dt \nonumber \\&=&\alpha K e^{\beta c} \beta^{p-1} \Gamma (1-p,\beta c), \nonumber
\end{eqnarray}
where $\Gamma(\cdot,\cdot)$ is the well-known Incomplete Gamma Function: $\Gamma (a,y) = \int_{y}^{\infty} t^{a-1} e^{-t} dt$.

\subsection{EXP x SQR}

For the combination ``EXPxSQR'', we have that, for stationarity to be achieved:
\begin{eqnarray}
0 \leq \int_{0}^{\infty} EXP(\alpha,\beta) SQR(B,L) dx < 1 \nonumber
\end{eqnarray}
\begin{eqnarray}
0 \leq \int_{0}^{L} \alpha B e^{-\beta x} dx < 1 \nonumber
\end{eqnarray}

Thus:
\begin{eqnarray}
\int_{0}^{L} \alpha B e^{-\beta x} dx = \left[ \dfrac{\alpha B e^{-\beta x}}{\beta} \right]_0^L = \dfrac{\alpha B (1- e^{-\beta L})}{\beta} \nonumber
\end{eqnarray}

So, in the case of a multiplicative combination, the SQR kernel acts as a truncation horizon.

\subsection{EXP x SNS}

For the combination ``EXPxSNS'', we have that, for stationarity to be achieved:
\begin{eqnarray}
0 \leq \int_{0}^{\infty} EXP(\alpha,\beta) SNS(A,\omega) dx < 1 \nonumber
\end{eqnarray}
\begin{eqnarray}
0 \leq \int_{0}^{\dfrac{\pi}{\omega}} A \alpha e^{-\beta x} sin (\omega x) dx < 1 \nonumber
\end{eqnarray}

Where:
\begin{eqnarray}
\int_{0}^{\dfrac{\pi}{\omega}} A \alpha e^{-\beta x} sin (\omega x) dx &=& \int_{0}^{\dfrac{\pi}{\omega}} A \alpha e^{-\beta x} \dfrac{e^{i \omega x} - e^{-i \omega x}}{2 i} dx \nonumber \\&=& \dfrac{A \alpha}{2 i} \left[ \dfrac{e^{(-\beta + i \omega)x}}{-\beta + i \omega} - \dfrac{e^{(-\beta - i \omega)x}}{-\beta - i \omega} \right]_0^{\dfrac{\pi}{\omega}} \nonumber \\&=& \dfrac{A \alpha}{2 i}  \left[ \dfrac{(-\beta -i \omega)e^{(-\beta + i \omega)x} - (-\beta +i \omega)e^{(-\beta - i \omega)x}}{\beta^2 + \omega^2} \right]_0^{\dfrac{\pi}{\omega}} \nonumber \\&=& \left[ \dfrac{A \alpha e^{-\beta x}}{2 i}  \dfrac{2 i \omega cos(\omega x) - 2 \beta sin(\omega x)}{\beta^2 + \omega^2} \right]_0^{\dfrac{\pi}{\omega}} \nonumber \\&=& \dfrac{A \alpha}{2 i} \dfrac{-2 i \omega (e^{\dfrac{- \beta \pi}{\omega}}-1)}{\beta^2+\omega^2} \nonumber \\&=& \frac{A \alpha \omega (1 + e^{\frac{-\beta \pi}{\omega}})}{(\omega^2 + \beta^2)} \nonumber
\end{eqnarray}

\subsection{PWL x PWL}

In the case of the combination ``PWLxPWL'', an upper bound is derived as follows:
\begin{eqnarray}
0 \leq \int_{0}^{\infty} PWL(K_1,c_1,p_1)PWL(K_2,c_2,p_2) dx <  1 \nonumber
\end{eqnarray}
\begin{eqnarray}
0 \leq \int_{0}^{\infty} \dfrac{K_{1}}{(x+c_{1})^{p_{1}}} \dfrac{K_{2}}{(x+c_{2})^{p_{2}}} dx < 1 \nonumber
\end{eqnarray}
Then:
\begin{eqnarray}
\int_{0}^{\infty} \dfrac{K_{1}}{(x+c_{1})^{p_{1}}} \dfrac{K_{2}}{(x+c_{2})^{p_{2}}} dx \nonumber  &\leq& \int_{0}^{\infty} \dfrac{K_{1} K_{2}}{(x+min(c_{1},c_{2}))^{p_{1}+p_{2}}} dx \nonumber \\&=& \dfrac{K_{1} K_{2}}{(p_{1}+p_{2}-1) min(c_{1},c_{2})^{(p_{1}+p_{2}-1)}} \nonumber
\end{eqnarray}

\subsection{PWL x SQR}

For the combination ``PWLxSQR'', we have that, for stationarity to be achieved:

\begin{eqnarray}
0 \leq \int_{0}^{\infty} PWL(K,c,p)SQR(B,L) dx < 1 \nonumber
\end{eqnarray}
\begin{eqnarray}
0 \leq \int_{0}^{L} \dfrac{KB}{(x+c)^{p}} dx < 1 \nonumber
\end{eqnarray}
Where:
\begin{eqnarray}
\int_{0}^{L} \dfrac{KB}{(x+c)^{p}} dx = \left[ \dfrac{KB}{(1-p)(x+c)^{(p-1)}}] \right]_0^L \nonumber = \frac{KB (c^{-(p-1)} - (c+L)^{-(p-1)})}{p-1} \nonumber
\end{eqnarray}

So, once again, the SQR kernel acts as a truncation horizon.

\subsection{PWL x SNS}

In the case of the combination ``PWLxSNS'', an upper bound is derived as follows:

\begin{eqnarray}
0 \leq \int_{0}^{\infty} PWL(K,c,p)SNS(A,\omega) dx < 1 \nonumber
\end{eqnarray}
\begin{eqnarray}
0 \leq \int_{0}^{\dfrac{\pi}{\omega}} \dfrac{KA sin(\omega x)}{(x+c)^{p}} dx < 1 \nonumber
\end{eqnarray}

Where:
\begin{eqnarray}
\int_{0}^{\dfrac{\pi}{\omega}} \dfrac{KA sin(\omega x)}{(x+c)^{p}} dx  &\leq& \int_{0}^{\dfrac{\pi}{\omega}} \dfrac{KA}{(x+c)^{p}} dx \nonumber \\&=& \left[ \dfrac{KA}{(1-p)(x+c)^{(p-1)}}] \right]_0^{\dfrac{\pi}{\omega}} \nonumber \\&=& KA\frac{((c + \frac{\pi}{\omega})^{1-p} - c^{1-p})}{1-p} \nonumber
\end{eqnarray}

\subsection{SQR x SQR}

For the combination ``SQRxSQR'', we have that, for stationarity to be achieved:
\begin{eqnarray}
0 \leq \int_{0}^{\infty} SQR(B_1,L_1)SQR(B_2,L_2) dx < 1 \nonumber
\end{eqnarray}
\begin{eqnarray}
0 \leq \int_{0}^{min(L_1,L_2)} B_1 B_2 dx < 1 \nonumber
\end{eqnarray}

Where:
\begin{eqnarray}
\int_{0}^{min(L_1,L_2)} B_1 B_2 dx = B_1 B_2 min(L_1,L_2) = B L \nonumber
\end{eqnarray}

So, the multiplicative combination of two SQR kernels may be reduced to the case of a single SQR kernel.

\subsection{SQR x SNS}

In the case of combinations of discontinuous kernels (SQR and SNS), we assume they have the same starting and ending points, i.e., $L = \dfrac{\pi}{\omega}$. So, for the combination ``SQRxSNS'', we have that, for stationarity to be achieved:

\begin{eqnarray}
0 \leq \int_{0}^{\infty} SQR(B,L)SNS(A,\omega) dx < 1 \nonumber
\end{eqnarray}
\begin{eqnarray}
0 \leq \int_{0}^{\dfrac{\pi}{\omega}} A B sin(\omega x) dx < 1 \nonumber
\end{eqnarray}

Where:
\begin{eqnarray}
\int_{0}^{\dfrac{\pi}{\omega}} A B sin(\omega x) dx = \dfrac{2AB}{\omega} \nonumber
\end{eqnarray}

\subsection{SNS x SNS}

In the case of combinations of discontinuous kernels (SQR and SNS), we assume they have the same starting and ending points. So, for the combination ``SNSxSNS'', we have that, for stationarity to be achieved:

\begin{eqnarray}
0 \leq \int_{0}^{\infty} SNS(A_1,\omega)SNS(A_2,\omega) dx < 1 \nonumber
\end{eqnarray}
\begin{eqnarray}
0 \leq \int_{0}^{\dfrac{\pi}{\omega}} A_1 A_2  sin^2 (\omega x) dx < 1 \nonumber
\end{eqnarray}

Where:
\begin{eqnarray}
\int_{0}^{\dfrac{\pi}{\omega}} A_1 A_2  sin^2 (\omega x) dx = \int_{0}^{\dfrac{\pi}{\omega}} A  \dfrac{(1-cos (2 \omega x))}{2} dx \nonumber = \dfrac{\pi A}{2 \omega} \nonumber
\end{eqnarray}

\section{Derivation of the Log-likelihood formula for HPs}

This derivation follows the steps on \cite{PL15}. Given a realization $(t_1,t_2,...,t_k)$ of some regular point process observed over the interval [0,T], the log-likelihood is expressed as:
\begin{eqnarray}
l = \sum_{i=1}^{k} \log (\lambda (t_i)) - \int_{0}^{T} \lambda (u) du \nonumber
\end{eqnarray}
\begin{proof}
Let be the joint probability density of the realization:
\begin{eqnarray}
L = f(t_1,t_2,...,t_k) = \prod_{i=1}^{k} f(t_i) \nonumber
\end{eqnarray}
It can be written in terms of the Conditional Intensity Function. We can then find f in terms of $\lambda$:
\begin{eqnarray}
\label{eq: apllh}
\lambda (t) = \dfrac{f(t)}{1 - F(t)} = \dfrac{\dv{F(t)}{t}}{1 - F(t)} = -\dv{\log (1-F(t))}{t}, \nonumber
\end{eqnarray}
where, given the history up to last arrival u, $\mathcal{H}(u)$, F(t) is then defined as the conditional cumulative probability distribution of the next arrival time $T_{k+1}$:
\begin{eqnarray}
F(t) = F(t|\mathcal{H}(u)) = \int_{u}^{t} f(s|\mathcal{H}(u)) ds \nonumber
\end{eqnarray}
Integrating both sides of Equation (\ref{eq: apllh}) over $(t_k,t)$:
\begin{eqnarray}
\label{eq: apllh2}
-\int_{t_k}^{t} \lambda (u) du = \log (1-F(t)) - \log (1-F(t_k)) \nonumber
\end{eqnarray}
Given that the realization is assumed to have come from a so-called \textit{simple process}, i.e., a process in which multiple arrivals cannot occur at the same time, we have that $F(t_k)$ = 0 as $T_{k+1} > t_k$, which simplifies equation (\ref{eq: apllh2}) to:
\begin{eqnarray}
-\int_{t_k}^{t} \lambda (u) du = \log (1-F(t)) \nonumber
\end{eqnarray}
Further rearranging the expression:
\begin{eqnarray}
F(t) = exp \left( -\int_{t_k}^{t} \lambda (u) du \right), \nonumber
\end{eqnarray}
and
\begin{eqnarray}
f(t) = \lambda (t) exp \left(-\int_{t_k}^{t} \lambda (u) du \right) \nonumber
\end{eqnarray}
Thus, the likelihood becomes:
\begin{eqnarray}
L = \prod_{i=1}^{k} f(t_i) = \prod_{i=1}^{k} \lambda (t_i) exp \left( -\int_{t_{i-1}}^{t_i} \lambda (u) du \right) \nonumber = \left[ \prod_{i=1}^{k} \lambda (t_i) \right] exp \left( -\int_{0}^{t_k} \lambda (u) du \right) \nonumber
\end{eqnarray}
Given that the process is observed on [0,T], the likelihood must include the probability of seeing no arrivals in $(t_k,T]$:
\begin{eqnarray}
L = \left[ \prod_{i=1}^{k}f(t_i)\right] (1 - F(T)) \nonumber
\end{eqnarray}
Through using the formulation of F(t), we have that:
\begin{eqnarray}
L = \left[ \prod_{i=1}^{k} \lambda (t_i)\right] exp \left( -\int_{0}^{T} \lambda (u) du\right) \nonumber
\end{eqnarray}
Finally, getting the logarithm of the expression, we have the formula for \textit{l}:
\begin{eqnarray}
l = \sum_{i=1}^{k} \log (\lambda (t_i)) - \int_{0}^{T} \lambda (u) du \nonumber
\end{eqnarray}
\end{proof}

\newpage
\section{Automatic Report}

\begin{figure}[H]
\centering
\includegraphics[width=0.5\linewidth]{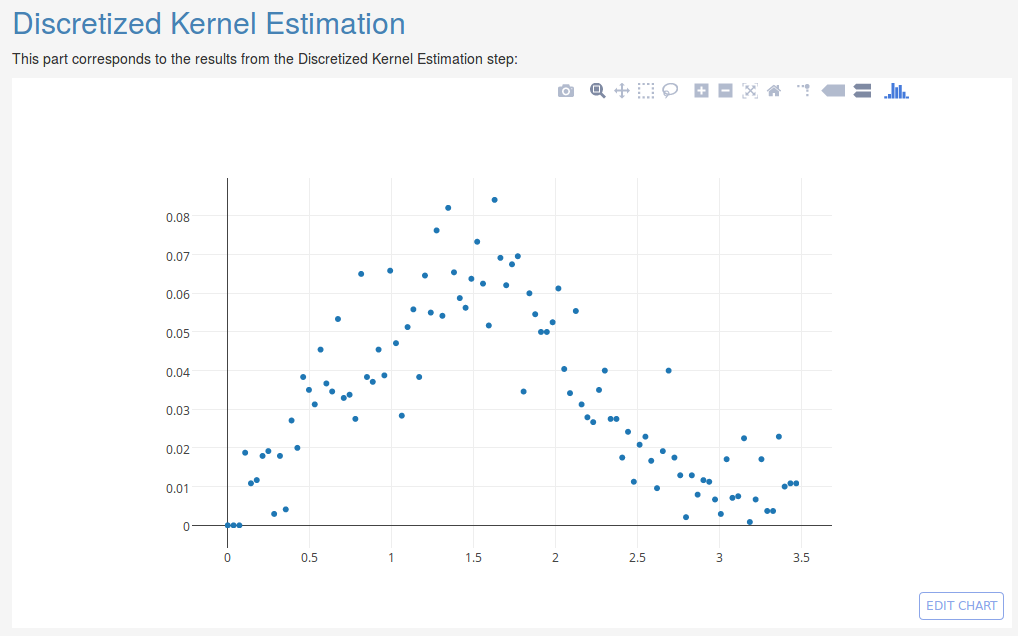}
\end{figure}
\begin{figure}[H]
\centering
\includegraphics[width=0.5\linewidth]{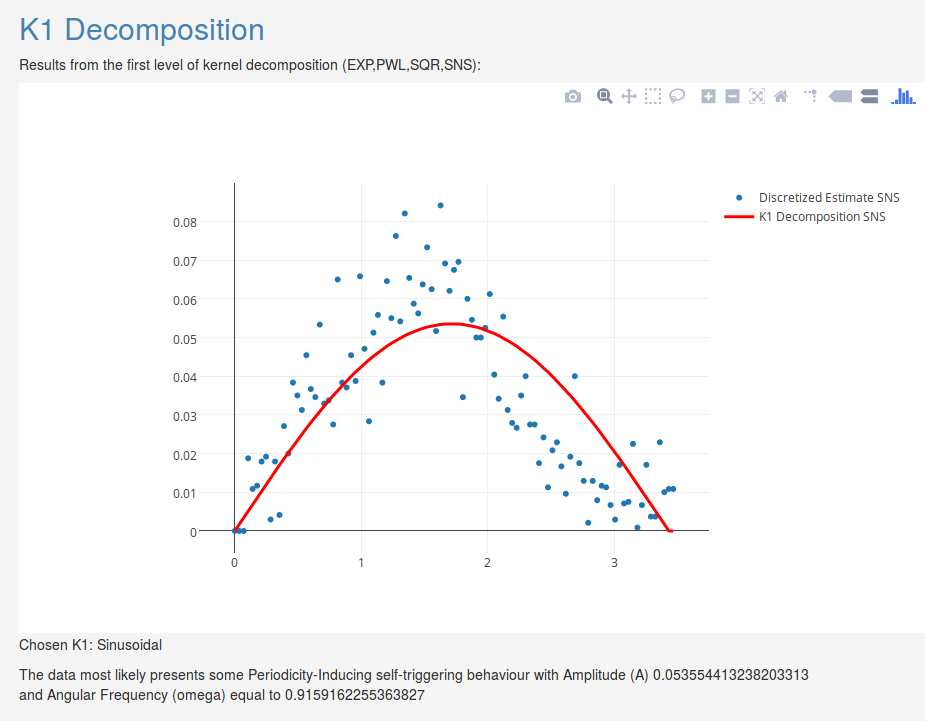}
\end{figure}
\begin{figure}[H]
\centering
\includegraphics[width=0.5\linewidth]{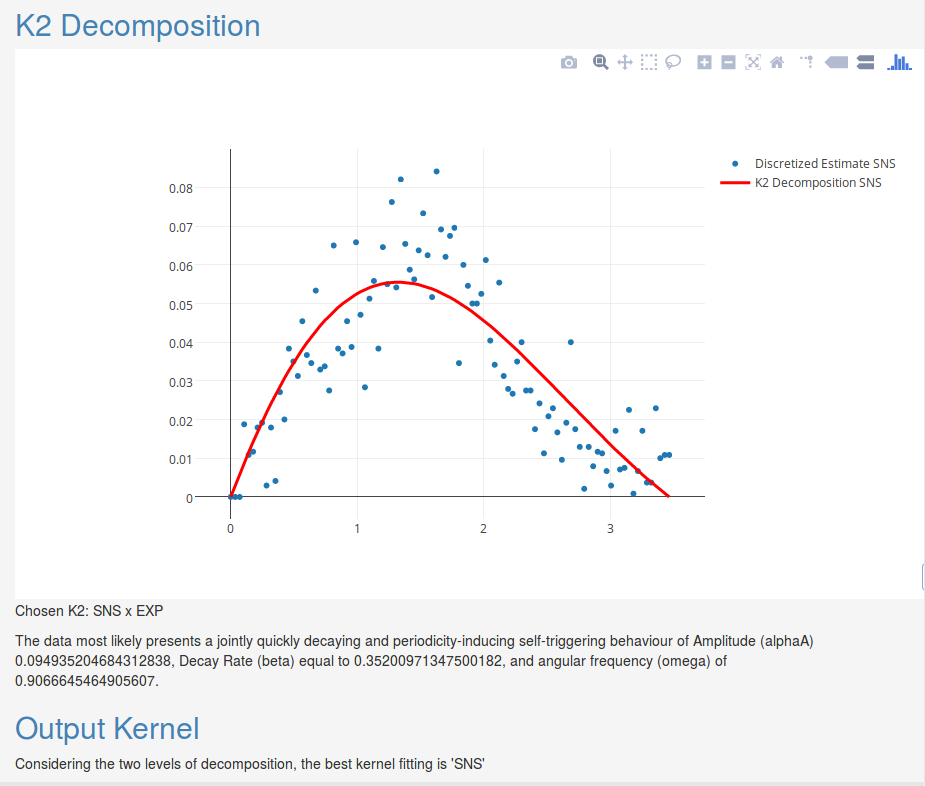}
\end{figure}

\section{Comparison between Gradient-based and Discretized Estimation steps for the financial datasets}

\begin{figure}[H] 
\centering
\includegraphics[width=\linewidth,height=2.0cm]{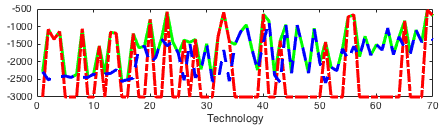} 
\includegraphics[width=\linewidth,height=2.0cm]{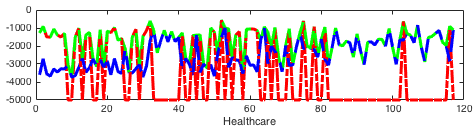} 
\includegraphics[width=\linewidth,height=2.0cm]{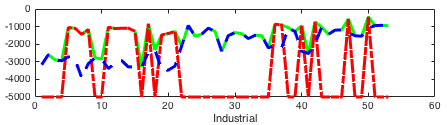} 
\includegraphics[width=\linewidth,height=2.0cm]{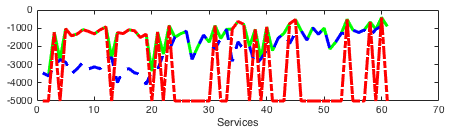} 
\includegraphics[width=\linewidth,height=2.0cm]{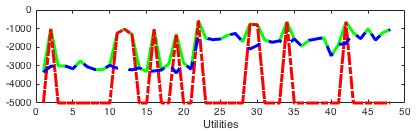} 
\caption{Comparison among loglikelihood of our kernel composition algorithm (Discretized Estimation) ({\color{blue} blue}), gradient descent exponential Hawkes ({\color{red} red}) and an Ensemble Model (Discretized Estimation + Gradient-based) ({\color{green} green}), for each valid sequence.}
\label{fig: comp_llh_GD} 
\end{figure}

\end{document}